# Motion Artifacts Correction from Single-Channel EEG and fNIRS Signals using Novel Wavelet Packet Decomposition in Combination with Canonical Correlation Analysis

Md Shafayet Hossain[1], Muhammad E. H. Chowdhury[2*], Mamun Bin Ibne Reaz[1*], Sawal H. M. Ali[1], Ahmad Ashrif A. Bakar[1], Serkan Kiranyaz[2], Amith Khandaker[2], Mohammed Alhatou[3], Rumana Habib[4], Muhammad Maqsud Hossain[5]

[1] Department of Electrical, Electronic and Systems Engineering, Universiti Kebangsaan Malaysia, Bangi, 43600, Selangor, Malaysia;
[2] Department of Electrical Engineering, Qatar University, Doha, 2713, Qatar;
[3] Neuromuscular Division, Hamad General Hospital and Department of Neurology; Alkhor Hospital, Doha, 3050, Qatar
[4] Department of Neurology, BIRDEM General Hospital, Dhaka-1000, Bangladesh
[5] NSU Genome Research Institute (NGRI), North South University, Dhaka-1229, Bangladesh
*Correspondence: Muhammad E.H. Chowdhury (mchowdhury@qu.edu.qa); Mamun Bin Ibne Reaz (mamun@ukm.edu.my)

**Abstract:** The electroencephalogram (EEG) and functional near-infrared spectroscopy (fNIRS) signals, highly non-stationary in nature, greatly suffers from motion artifacts while recorded using wearable sensors. Since successful detection of various neurological and neuromuscular disorders is greatly dependent upon clean EEG and fNIRS signals, it is a matter of utmost importance to remove/reduce motion artifacts from EEG and fNIRS signals using reliable and robust methods. In this regard, this paper proposes two robust methods: i) Wavelet packet decomposition (WPD), and ii) WPD in combination with canonical correlation analysis (WPD-CCA), for motion artifact correction from single-channel EEG and fNIRS signals. The efficacy of these proposed techniques is tested using a benchmark dataset and the performance of the proposed methods is measured using two well-established performance matrices: i) Difference in the signal to noise ratio ($\Delta SNR$) and ii) Percentage reduction in motion artifacts ($\eta$). The proposed WPD-based single-stage motion artifacts correction technique produces the highest average $\Delta SNR$ (29.44 dB) when db2 wavelet packet is incorporated whereas the greatest average $\eta$ (53.48%) is obtained using db1 wavelet packet for all the available 23 EEG recordings. Our proposed two-stage motion artifacts correction technique i.e. the WPD-CCA method utilizing db1 wavelet packet has shown the best denoising performance producing an average $\Delta SNR$ and $\eta$ values of 30.76 dB and 59.51%, respectively for all the EEG recordings. On the other hand, for the available 16 fNIRS recordings, the two-stage motion artifacts removal technique i.e. WPD-CCA has produced the best average $\Delta SNR$ (16.55 dB, utilizing db1 wavelet packet) and largest average $\eta$ (41.40%, using fk8 wavelet packet). The highest average $\Delta SNR$ and $\eta$ using single-stage artifacts removal techniques (WPD) are found as 16.11 dB and 26.40%, respectively for all the fNIRS signals using fk4 wavelet packet. In both EEG and fNIRS modalities, the percentage reduction in motion artifacts increases by 11.28% and 56.82%, respectively when two-stage WPD-CCA techniques are employed in comparison with the single-stage WPD method. In addition, the average $\Delta SNR$ also increases when WPD-CCA techniques are used instead of single-stage WPD for both EEG and fNIRS signals. The increment in both $\Delta SNR$ and $\eta$ values is a clear indication that two-stage WPD-CCA performs relatively better compared to single-stage WPD. The results reported using the proposed methods outperform most of the existing state-of-the-art techniques.

## 1. Introduction

Due to the paradigm shift of hospital-based treatment in the direction of wearable and ubiquitous monitoring, nowadays, the acquisition and processing of vital physiological signals have become prevalent in the ambulatory setting. Since the acquisition of physiological signals is inclined to movement artifacts that happen due to the deliberate and/or voluntary movement of the patient during signal procurement utilizing wearable devices, restricting patients totally from physical movements, intentional and/or unintentional, is exceptionally troublesome. As a result, the physiological signals may get corrupted to some degree by motion artifacts. In some instances, this defilement may end up so conspicuous that the recorded signals may lose their usability unless the movement artifacts are diminished significantly.

Electroencephalogram (EEG) measures the electrical activity of the human brain quantitatively which took place due to the firing of neurons [1] and such brain activity is recorded utilizing a good number of cathodes which are located at different regions of the scalp [2]. EEG is one of the key diagnostic tests for epileptic seizure detection [3, 4]. Other decisive utilization of EEG includes the estimation of drowsiness levels [5-8], emotion detection [9], cognitive workload [6, 10], and brain-computer interfaces (BCIs) [11-16]. All of which have potential applications in the personal healthcare domain. Lately, the implementation of EEG-based biometric systems utilizing the inborn anti-spoofing capability of EEG signals was studied and appeared to be promising [17].

The functional near-infrared spectroscopy (fNIRS), a non-invasive optical brain imaging technique, measures changes in hemoglobin (Hb) concentrations inside the human brain [18] by employing light of various wavelengths in the infrared band and estimating the difference in the optical absorption [19]. Medical applications of fNIRS mainly focus on the noninvasive measurement of brain functions [20, 21], cognitive tasks identification [22, 23], and BCI [24-26].

Apart from movement artifacts, physiological signals undergo other types of artifacts as well. Gradient artifacts (GA) and pulse artifacts (PA) are the two most frequent artifacts observed in EEG during the simultaneous EEG-fMRI tests [27-29]. On the other hand, event-related fNIRS signals are regularly sullied by heartbeat, breath, Mayer waves, etc., as well as extra-cortical physiological clamors from the superficial layers [30].

To reduce motion artifacts from EEG, numerous endeavor was made previously and were summarized in [31, 32]. In [33], the performance of motion artifacts correction techniques utilizing discrete wavelet transform (DWT) [34], empirical mode decomposition (EMD) [35], ensemble empirical mode decomposition (EEMD) [36], EMD along with canonical correlation analysis (EMD-CCA), EMD with independent component analysis (EMD-ICA), EEMD with ICA (EEMD-ICA), and EEMD with CCA (EEMD-CCA) were reported. Maddirala and Shaik [37] used singular spectrum analysis (SSA) [38] whereas DWT along with the thresholding technique was utilized in [39]. Gajbhiye et al. [40] employed wavelet-based transform along with the total variation (TV) and weighted TV (WTV) denoising techniques whereas in [41], wavelet domain optimized Savitzky–Golay filter was proposed for the removal of motion artifacts from EEG. . Recently, Hossain et. al. [42] utilized variational mode decomposition (VMD) [43] for the correction of motion artifacts from EEG signals.

In the last few decades, multiple motion artifacts removal techniques was proposed [44-46] for the removal of motion artifacts from the fNIRS signal. Sweeney et al. [47] used adaptive filter, Kalman Filter, and EEMD-ICA. Scholkmann et al. [48] utilized the moving standard deviation and spline interpolation method whereas in [49] wavelet-based method was proposed. The authors of [33] used DWT, EMD, EEMD, EMD-ICA, EEMD-ICA, EMD-CCA, and EEMD-CCA. In [50], Barker et al. used an autoregressive model-based algorithm whilst kurtosis-based wavelet transform was proposed in [51], and Siddiquee et al. [52] utilized nine-degree of freedom inertia measurement unit (IMU) data to mathematically estimate the movement artifacts in the fNIRS signal using autoregressive exogenous (ARX) input model. A hybrid algorithm was proposed in [53] to filter out the movement artifacts from fNIRS signals where both the spline interpolation method and Savitzky–Golay filtering were employed. Very recently, the two-stage VMD-CCA technique was employed in [42].

The development of robust algorithms that can successfully reduce motion artifacts significantly from EEG and fNIRS data is critical; otherwise, the signals' interpretation could be erroneous by medical doctors and/or machine-learning-based applications. As mentioned earlier, DWT, EMD, EEMD, VMD, DWT-ICA, EMD-ICA, EEMD-ICA, EMD-CCA, EEMD-CCA, VMD-CCA, etc. were the most commonly used methods for the correction of motion artifacts from EEG and fNIRS signals. ICA and CCA can not be used independently for single-channel EEG/fNIRS motion artifacts correction as the input of ICA/CCA algorithms require at least two (or more) channels data whereas DWT, EMD, EEMD, VMD, etc. algorithms suffer from several limitations which are discussed in the discussion section of this paper. Also, there is still room for improvement for $\Delta SNR$ and $\eta$ values which can be achieved using other effective novel methods. Therefore, in this paper, two novel motion artifacts removal techniques have been proposed which can eliminate motion artifacts from single-channel EEG and fNIRS signals to a great extent. The first is a single-stage motion artifacts correction technique using the wavelet packet decomposition (WPD) whereas the other novel method is WPD in combination with CCA (WPD-CCA), a two-stage motion artifacts removal technique, as the name suggests.

In this extensive study, for the correction of motion artifact from EEG and fNIRS signals using the WPD method, four different wavelet packet families (Daubechies (db*N*), Symlets (sym*N*), Coiflets (coif*N*), Fejer-Korovkin (fk*N*)) have been used with three different vanishing moments (for each of the wavelet packet) that resulted in a total of 12 different investigations. The wavelet packets used in the WPD method are db1, db2, db3, sym4, sym5, sym6, coif1, coif2, coif3, fk4, fk6, and fk8. To the best of our knowledge, the WPD algorithm has not been used for the removal of motion artifacts from single-channel EEG and fNIRS signals to date. WPD-CCA method is another novel contribution of this research work where Daubechies and Fejer-Korovkin wavelet packet families are utilized. In the WPD-CCA technique, db1, db2, db3, fk4, fk6, and fk8 have been used separately resulting in 6 different investigations to reduce motion artifacts from EEG and fNIRS signals more efficiently.

The rest of this paper is organized as follows: Section II discusses the theoretical background of the different algorithms (WPD, CCA, WPD-CCA) investigated here, Section III provides brief information about the EEG and fNIRS benchmark dataset, and experimental methodology. Section IV provides the results of the artifact removal techniques proposed in this work and section VI covers the discussion. Finally, the paper is concluded in section V.

## 2. Theoretical background
### 2.1 Wavelet packet decomposition (WPD)

Using the WPD technique, signals can be decomposed into a wavelet packet basis at diverse scales [54, 55]. For $j$-level decomposition, a wavelet packet basis is represented by multiple signals $[(n - 2^j k)]_{k \in \mathbb{Z}}$ where $i \in \mathbb{Z}^+$, $0 \leq i \leq 2j - 1$. The wavelet packet bases $\psi_j^i(n)$, are produced recursively from the scaling and wavelet functions, $\psi_1^0(n) = \phi(n)$ and $\psi_1^1(n) = \psi(n)$ respectively, as follows:

$$\psi_j^{2i}(n) = \sum_k h(k)\psi_{j-1}^i(n - 2^{j-1}k) \quad (1)$$

$$\psi_j^{2i+1}(n) = \sum_k g(k)\psi_{j-1}^i(n - 2^{j-1}k) \quad (2)$$

where $h(n)$ represents lowpass filter and $g(n)$ is the highpass filter defined as [54, 56]:

$$h(k) = \langle \psi_j^{2i}(u), \psi_{j-1}^i(u - 2^{j-1}k) \rangle \quad (3)$$

$$g(k) = \langle \psi_j^{2i+1}(u), \psi_{j-1}^i(u - 2^{j-1}k) \rangle \quad (4)$$

The decomposition of a signal $x(n)$ onto the wavelet basis $j(n)$ at level $j$ can be expressed as:

$$x(n) = \sum_{i,k} X_j^i \psi_j^i(n - 2^j k) \quad (5)$$

where $X_j^i(k)$ signifies the $k$th wavelet coefficient of the packet $i$, at level $j$. Here, $X_j^i(k)$ represents the intensity of the localized wavelet $\psi_j^i(n - 2^j k)$, defined by:

$$X_j^i(k) = \langle x(n), \psi_j^i(n - 2^j k) \rangle \quad (6)$$

Let $x(n)$ represent a recorded EEG/fNIRS signal which can be expressed as the sum of a source signal $s(n)$ and a motion artifact signal $v(n)$ as follows:

$$x(n) = s(n) + v(n) \quad (7)$$

In general, the source signal $s(n)$ is assumed to be normally distributed having a mean value equals to zero, $s(n) \sim N(0, \sigma)$, where $\sigma^2$ characterizes the variance of $s(n)$ [57]. On the other hand, general assumptions regarding the artifact signal $v(n)$ includes temporal localization, not normally distributed with high local variance.

According to [58], $X_j^i(k)$, can be represented as the sum of $S_j^i(k)$ and $V_j^i(k)$ where $X_j^i(k)$, $S_j^i(k)$, and $V_j^i(k)$ are the wavelet coefficients of $x(n)$, $s(n)$, and $v(n)$, respectively:

$$X_j^i(k) = S_j^i(k) + V_j^i(k) \quad (8)$$

It is noteworthy to mention that the wavelet coefficients $V_j^i(k)$ will be sparse as well as the non-zero coefficients will have a relatively higher magnitude as the variance of $v(n)$ is locally high which would cause an increase in the local variance of the recorded EEG/fNIRS signal $x(n)$.

### 2.2 Canonical Correlation Analysis (CCA)

CCA [59] is one of the most popular blind source separation methods which has the capability of dissociating multiple mixed or noisy signals. Assuming linear mixing, square mixing, and stationary mixing [60], the CCA technique computes an un-mixing matrix **W** which helps identify the unknown independent components **Ŝ** from a matrix **X** which is a recorded multi-channel signal as follows:

$$\hat{\mathbf{S}} = \mathbf{WX} \tag{9}$$

CCA also estimates the unknown independent components $\hat{\mathbf{S}}$ using Equation 9 utilizing second-order statistics (SOS). CCA forcefully makes the sources to be autocorrelated maximally as well as makes the sources mutually uncorrelated [61]. Let us assume $\mathbf{y}$ as a linear combination of neighboring samples for an input signal $\mathbf{x}$ (i.e. $y(t) = x(t-1) + x(t+1)$) [62]. Consider the linear combinations of the components in $\mathbf{x}$ and $\mathbf{y}$, known as the the canonical variates,

$$x = \mathbf{w}_x^T(\mathbf{x} - \bar{\mathbf{x}}) \tag{10}$$
$$y = \mathbf{w}_y^T(\mathbf{y} - \bar{\mathbf{y}}) \tag{11}$$

where $\mathbf{w}_x$ and $\mathbf{w}_y$ represents the weight matrices. CCA computes $\mathbf{w}_x$ and $\mathbf{w}_y$ in such a way so that the correlation $\rho$ between $x$ and $y$ will be maximized [62]:

$$\rho = \frac{\mathbf{w}_x^T \mathbf{C}_{xy} \mathbf{w}_y^T}{\sqrt{\mathbf{w}_x^T \mathbf{C}_{xx} \mathbf{w}_x \mathbf{w}_y^T \mathbf{C}_{yy} \mathbf{w}_y}} \tag{12}$$

where $\mathbf{C}_{xx}$ and $\mathbf{C}_{yy}$ signify the nonsingular within-set covariance matrices and $\mathbf{C}_{xy}$ represent the between-sets covariance matrix. The maximized $\rho$ is calculated by setting the derivatives of Equation 12 (with respect to $\mathbf{w}_x$ and $\mathbf{w}_y$) equals to zero.

$$\mathbf{C}_{xx}^{-1} \mathbf{C}_{xy} \mathbf{C}_{yy}^{-1} \mathbf{C}_{yx}^T \hat{\mathbf{w}}_x = \rho^2 \hat{\mathbf{w}}_x$$
$$\mathbf{C}_{yy}^{-1} \mathbf{C}_{yx} \mathbf{C}_{xx}^{-1} \mathbf{C}_{xy}^T \hat{\mathbf{w}}_y = \rho^2 \hat{\mathbf{w}}_y \tag{13}$$

$\mathbf{w}_x$ and $\mathbf{w}_y$ can then be found out as the eigenvectors of the matrices $\mathbf{C}_{xx}^{-1} \mathbf{C}_{xy} \mathbf{C}_{yy}^{-1} \mathbf{C}_{yx}^T$ and $\mathbf{C}_{yy}^{-1} \mathbf{C}_{yx} \mathbf{C}_{xx}^{-1} \mathbf{C}_{xy}^T$ respectively and the corresponding eigenvalues $\rho^2$ are the squared canonical correlations. It is sufficient to solve only one of the eigenvalue equations to obtain the un-mixing matrix $\mathbf{W}$ as the solutions are related. Further, the underlying source signals $\hat{\mathbf{S}}$ can be estimated.

The components that seem to be artifacts can then be discarded by simply setting the corresponding columns of the $\hat{\mathbf{S}}$ matrix to zero before the signal reconstruction.

**2.3 WPD-CCA:**

The WPD algorithm can be utilized to decompose a single-channel signal into multi-channel signal $\mathbf{X}$ where each column of matrix $\mathbf{X}$ represents the detailed and approximated sub-band signals. The total number of generated sub-band signals would be equal to $2^j$ where $j$ denotes the level, *a priori*. To estimate the underlying true sources $\hat{\mathbf{S}}$ (Equation 9), these generated sub-band signals can then be used as the multi-channel input signals to the CCA algorithm. After that, the component/s of $\hat{\mathbf{S}}$ which seem to be artifacts can be discarded by making the corresponding columns of the matrix $\hat{\mathbf{S}}$ equal to zero. Bypassing this newly obtained source matrix through the inverse of the un-mixing matrix $\mathbf{W}^{-1}$, the multi-channel signals $\hat{\mathbf{X}}$ can be obtained. Finally, the cleaner signal $\hat{x}$ can be produced by simply summing all the columns of the matrix $\hat{\mathbf{X}}$.

**3. Methods**

This section describes the benchmark dataset used, pre-processing, study design, motion component identification, and evaluation metrics.

**3.1 Dataset Description**

A publicly available PhysioNet dataset [32, 33, 63] is used in this study that contains "reference ground truth" and motion corrupted signals for both EEG and fNIRS modalities. The details of the data recording procedure for EEG and fNIRS modalities were mentioned in [47]. During the data acquisition, two channels having the same hardware properties were placed on the test subject's scalp at very close proximity (20 mm for EEG modality and 30 mm for fNIRS modality) where the first channel was impacted with motion artifacts for 10-25 seconds at regular 2 minutes interval and the second channel was left untouched and undisturbed for the entire recording period. From the unimpacted channel (2nd channel), EEG/fNIRS signal was extracted which was free from motion artifacts and referred to as "reference ground truth" signal whereas the impacted channel (1st channel) provided EEG/fNIRS signal corrupted with motion artifacts. It is worthwhile to mention that both the motion corrupted and "reference ground truth" signals were extracted simultaneously from channels 1 and 2 respectively for approximately 9 minutes for each of the trial/test subjects. Also, the same channels were used to extract EEG/fNIRS data from all of the test subjects.

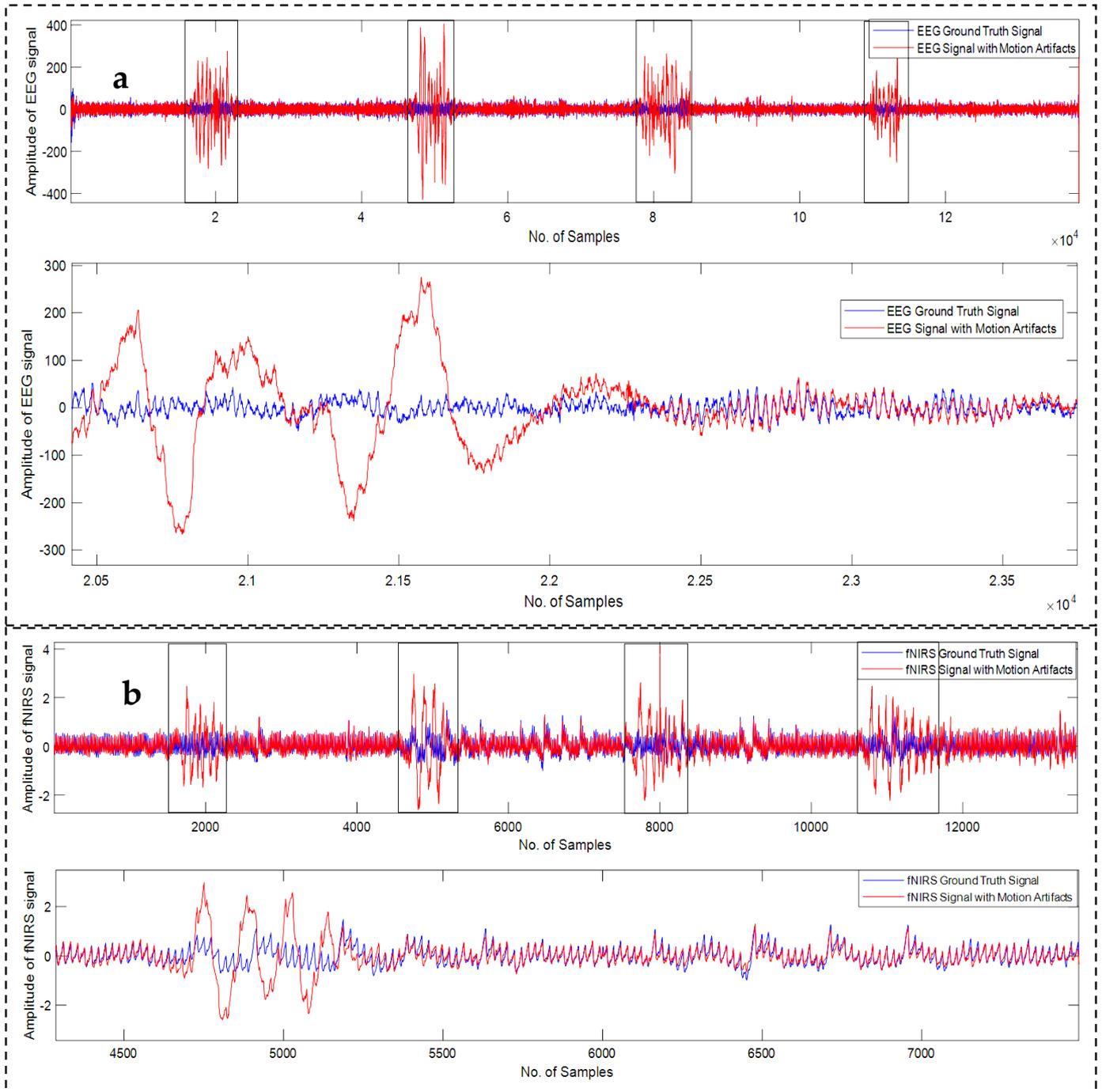

Figure 1: Example of motion-corrupted EEG (a) and fNIRS (b) signals. Two signals (blue: ground truth and red: motion-corrupted) are highly correlated during the motion artifacts free epochs. Boxed areas show the epochs of motion corrupted signals. A zoomed version is presented underneath each sub-plot.

Twenty-three sets of EEG recordings, sampled at 2048 Hz, collected from six patients in four different sessions are available in the database. Each recording consists of one motion corrupted EEG signal and one reference "ground truth" EEG signal. The average correlation coefficient between the reference "ground truth" and motion corrupted EEG signals is very high over the epochs where the motion artifacts are absent and the average correlation coefficient drops significantly during the epochs of motion artifacts [32]. The superimposed reference "ground truth" and motion corrupted EEG signals are illustrated in Figure 1a.

fNIRS signals were recorded at two different wavelengths: 690 nm and 830 nm wavelengths. There were 16 sets of fNIRS recordings (9 recordings at 830 nm wavelength and 7 recordings at 690 nm wavelength) in total from 10 test subjects at a sampling frequency of 25 Hz [33, 63]. Like EEG recordings, each recording of fNIRS consists of one motion corrupted fNIRS signal and one "reference ground truth" fNIRS signal. The overlaid "reference ground truth" fNIRS signal and motion artifact contaminated fNIRS signal is depicted in Figure 1b.

**3.2 Signal Preprocessing**

<u>Downsampling</u>: As EEG signals can be partitioned into a few sub-bands, specifically delta (1–4 Hz), theta (4–8 Hz), alpha (8–13 Hz), beta (13–30 Hz), and gamma (30–80 Hz) [64], we downsampled all the 23 sets of EEG recordings from 2048 Hz to 256 Hz which guarantees data reliability without losing any vital signal information and morphology. The fNIRS signals were not upsampled/downsampled as the original sampling rate was 25 Hz during acquisition.

<u>Power line noise removal:</u> To remove power line noise, a 3$^{rd}$ order Butterworth notch filter with a center frequency of 50 Hz was utilized to remove 50 Hz and its subsequent harmonics as a pre-processing technique for all the EEG and fNIRS signals.

<u>Baseline Drift Correction</u>: Both the EEG and fNIRS signals were found to have significant baseline drift, which is defined as undesired amplitude shifts in the signal that would result in inaccurate results if not corrected. To remove baseline drift from EEG and fNIRS recordings, a polynomial curve fitting method was used to estimate the baseline, which was then subtracted from the recorded raw signal.

**3.3 Study Design**

The simulations of this work were carried out in a PC with Intel(R) Core(TM) i5-8250U CPU at 1.80GHz which was equipped with 8 GB RAM. In-house-built MATLAB code was written to pre-process the EEG and fNIRS data. The single-stage WPD and two stages WPD-CCA methods were deployed in "MATLAB R2020a, The MathWorks, Inc". Figure 2 depicts the motion artifacts elimination framework presented in this study. An automated way for identifying motion corrupted components of the preprocessed signal is also discussed.

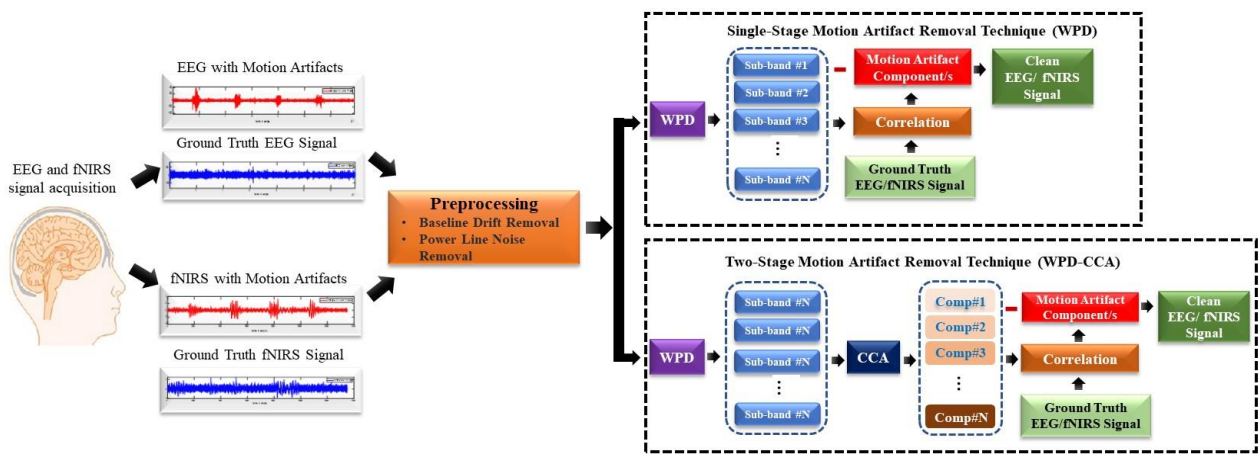

Figure 2: Methodological framework for the motion artifact correction.

In this study, the whole 9 minutes of EEG/fNIRS data of each trial were analyzed at one time using WPD and WPD-CCA methods. As mentioned earlier, WPD generates $2^j$ number of sub-band signals where level, $j$ is user-defined. Choosing j = 3 would produce 8 sub-band components where the probability of getting mixed of motion-corrupted components and artifacts-free signal components would be very high. Also, j=5 would produce 32 sub-band signals which would increase the computational complexity of the algorithm. Hence, In this research work, we have chosen $j$ equal to 4 for both EEG and fNIRS recordings that produced 16 sub-band signals/components in total for each of the EEG/fNIRS signals and ensured optimum performance. Again 12 different wavelet packets (db1, db2, db3, sym4, sym5, sym6, coif1, coif2, coif3, fk4, fk6, and fk8) were used in the single-stage motion artifact correction technique i.e. WPD. Among these 12 wavelet packets, 6 wavelet packets (db1, db2, db3, fk4, fk6, and fk8) were used in the WPD-CCA method due to the relatively better performance shown by Daubechies and Fejer-Korovkin wavelet packet families incorporated in the WPD technique. As several wavelet packets were used in this study, in the rest of the manuscript, a subscript is added with WPD to denote the corresponding wavelet packet used. As an example, $WPD_{(db1)}$ would refer to that the db1 wavelet packet is used.

With the availability of sub-band signals decomposed using the WPD technique, the artifact components can then be selected and removed. All the remaining sub-band signals can then either be added up to reconstruct a cleaner signal or all the sub-band signals can be fed as inputs to the CCA algorithm to determine the motion corrupted components to enhance the signal quality further.

CCA technique needs the number of input channels to be at least two or greater. In this work, single-channel EEG and fNIRS signals have been evaluated for the correction of motion artifacts. Hence, it is required to generate several sub-band signals which would be used as the inputs for the CCA algorithm. Six different WPD-CCA-based ($WPD_{(db1)}$-CCA, $WPD_{(db2)}$-CCA, $WPD_{(db3)}$-CCA, $WPD_{(fk4)}$-CCA, $WPD_{(fk6)}$-CCA, and $WPD_{(fk8)}$-CCA) two-stage artifacts removal technique has been realized for both single-channel EEG and fNIRS signals.

**3.4 Removal of Motion Artifact Components using "Reference Ground Truth" Method:**

A common challenge in eliminating motion artifacts utilizing the aforementioned artifact removal approaches is consistently identifying and removing the motion corrupted components from the signal of interest and reconstructing a cleaner signal. The available reference "ground truth" signal of EEG and fNIRS modalities were used to identify the motion corrupted components as well as test the efficacy of the proposed algorithms. If a component of the decomposed signal is removed and the signal is rebuilt using the other components, the correlation coefficient between the newly reconstructed signal and the ground truth signal will only rise if the removed component has motion artifacts. Using this basic yet efficient notion, motion artifact-affected components of the decomposed signal were discovered and discarded to reconstruct a cleaner signal, ensuring the best performance of each suggested technique during evaluation.

Figure 3a shows an example motion corrupted EEG signal and below Figure 3b represents the corresponding 16 sub-band components generated from that corresponding EEG signal using $WPD_{(sym4)}$ algorithm. Figure 4a depicts an example motion corrupted EEG signal and Figure 4b represents the resultant 16 CCA components where the input of the CCA method was 16 sub-band signals generated from the motion corrupted EEG signal using $WPD_{(coif1)}$.

Similarly, Figure 5a and Figure 6a show 2 separate motion corrupted fNIRS signals whereas *Figure 5*b, and *Figure 6*b represent the sub-band signals generated from $WPD_{(db1)}$, and 16 output CCA components where the input of the CCA algorithm was 16 sub-band signals generated from the motion corrupted EEG signal using $WPD_{(fk8)}$, respectively.

From visual inspection of the components generated from the single-stage (WPD) and two-stage (WPD-CCA) motion artifacts removal techniques, it can be stated that in most of the cases, motion artifacts components are usually found in one or two approximation sub-band/CCA components. Although this was the case for most of the EEG and fNIRS recordings, rather than blindly discarding these one or two sub-band/CCA components as motion artifact components, only those components were discarded, when removed, improved the correlation coefficient of the reconstructed signal in comparison with the available reference "ground truth" signal.

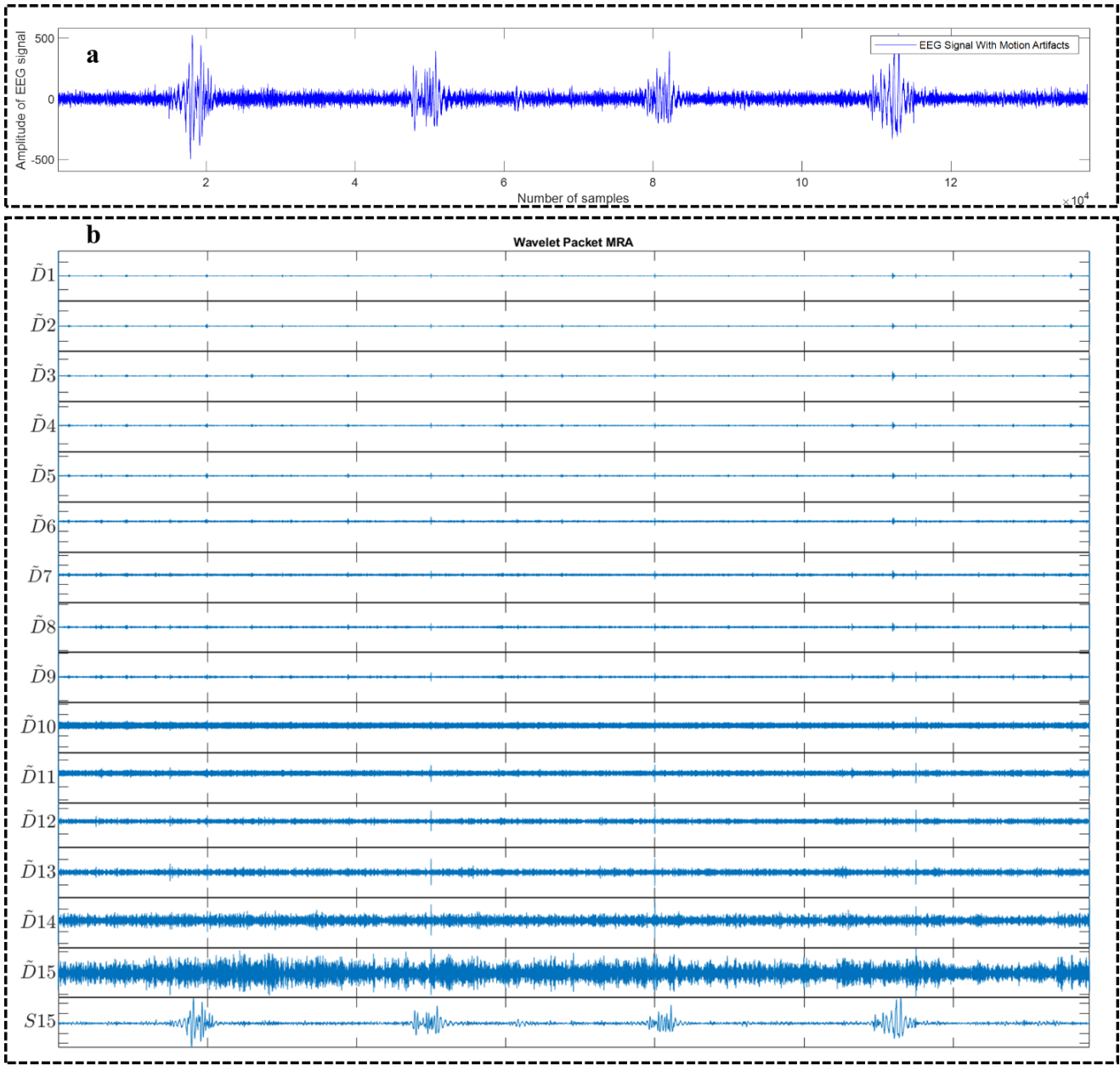

Figure 3: An example motion-corrupted single-channel EEG signal (a) and corresponding 16 sub-band components generated using WPD$_{(sym4)}$ algorithm (b). S15 denotes the Approximation sub-band signal having the lowest center frequency compared to the other sub-band signals i.e. D1-D15.

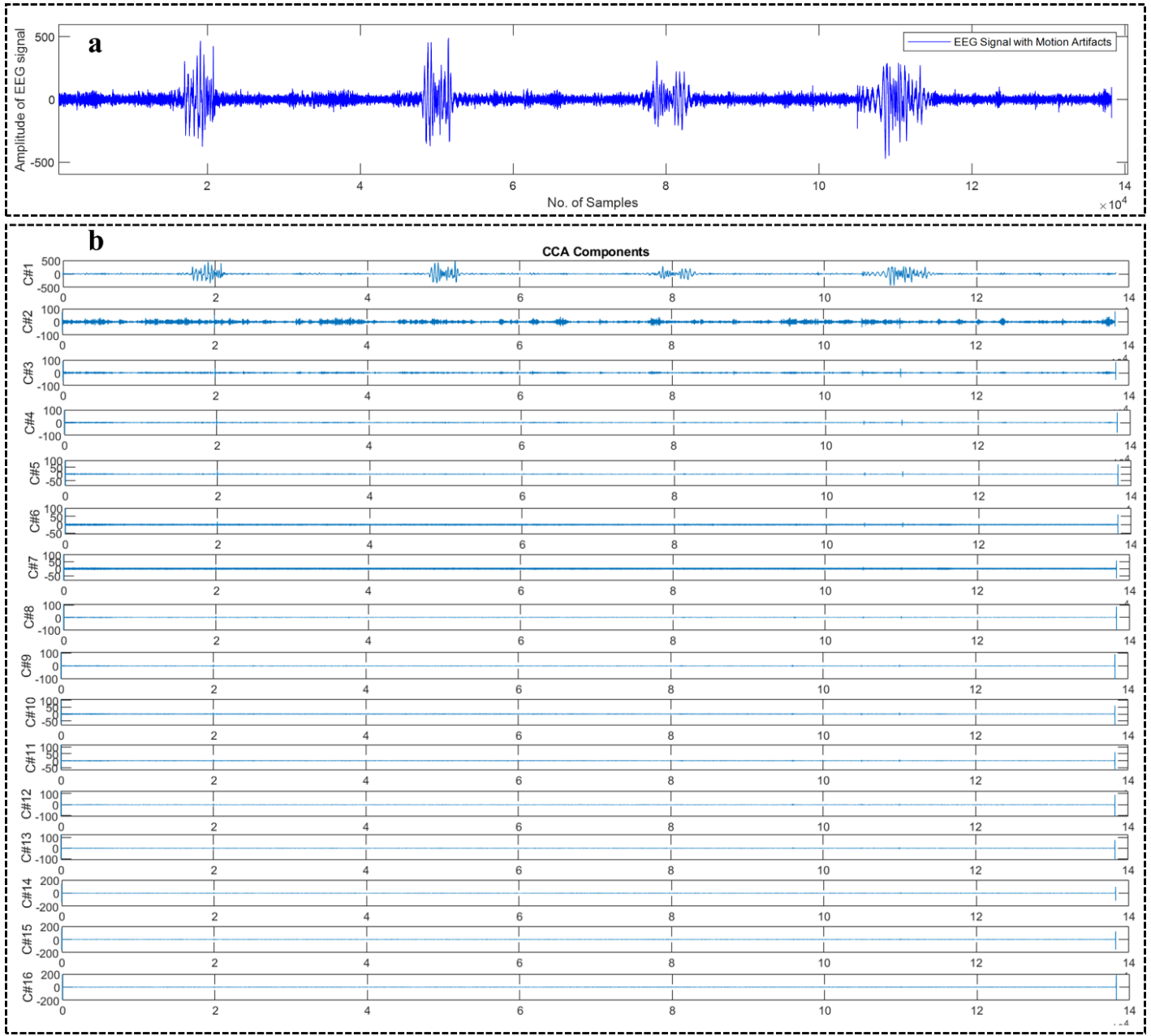

Figure 4: An example motion-corrupted single-channel EEG signal (a) and corresponding 16 CCA components generated from CCA algorithm (b)

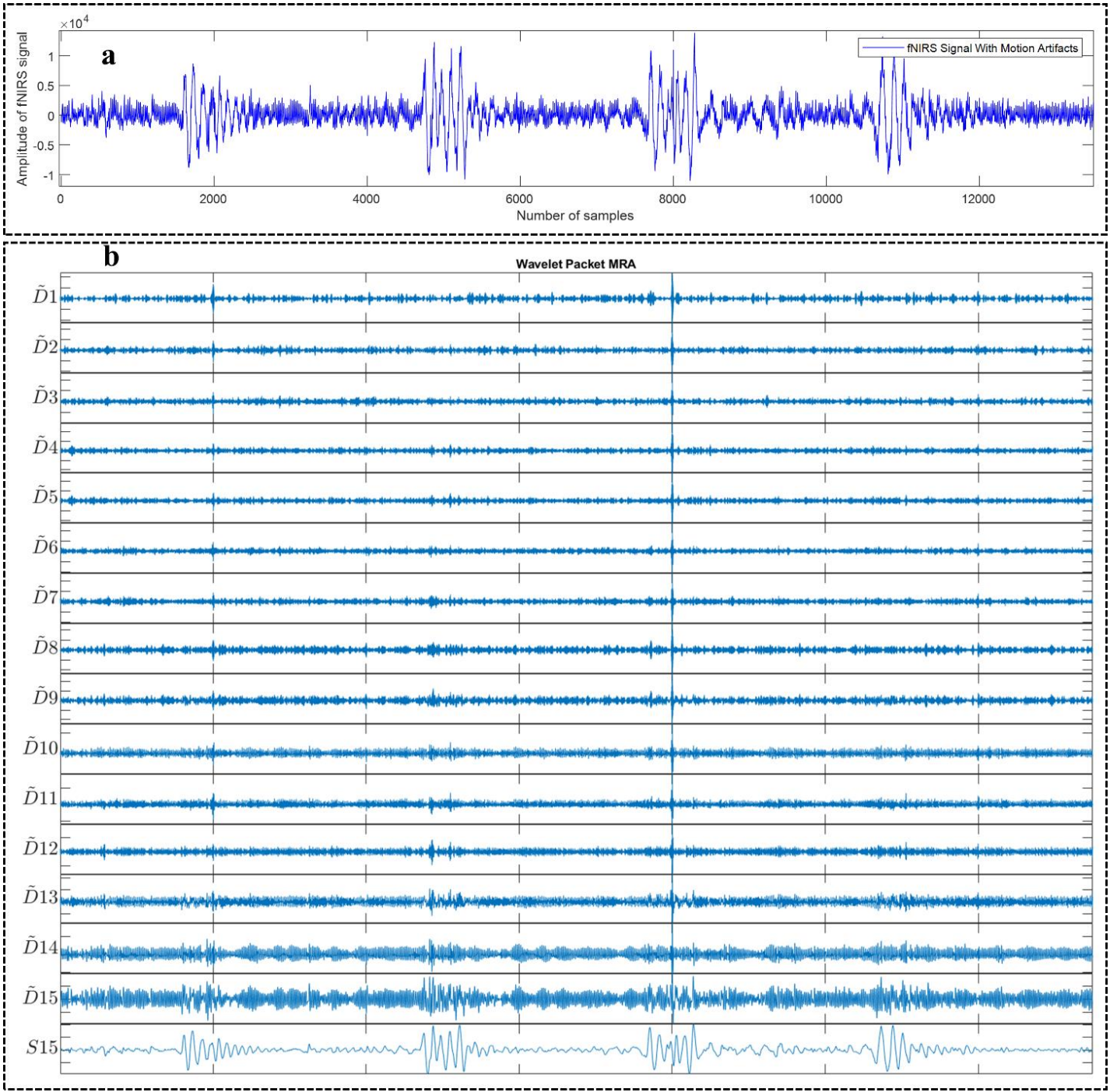

Figure 5: An example motion-corrupted single-channel fNIRS signal (a) and corresponding 16 sub-band components generated using WPD(db1) algorithm (b). S15 denotes the Approximation sub-band signal having the lowest center frequency compared to the other sub-band signals i.e. D1-D15.

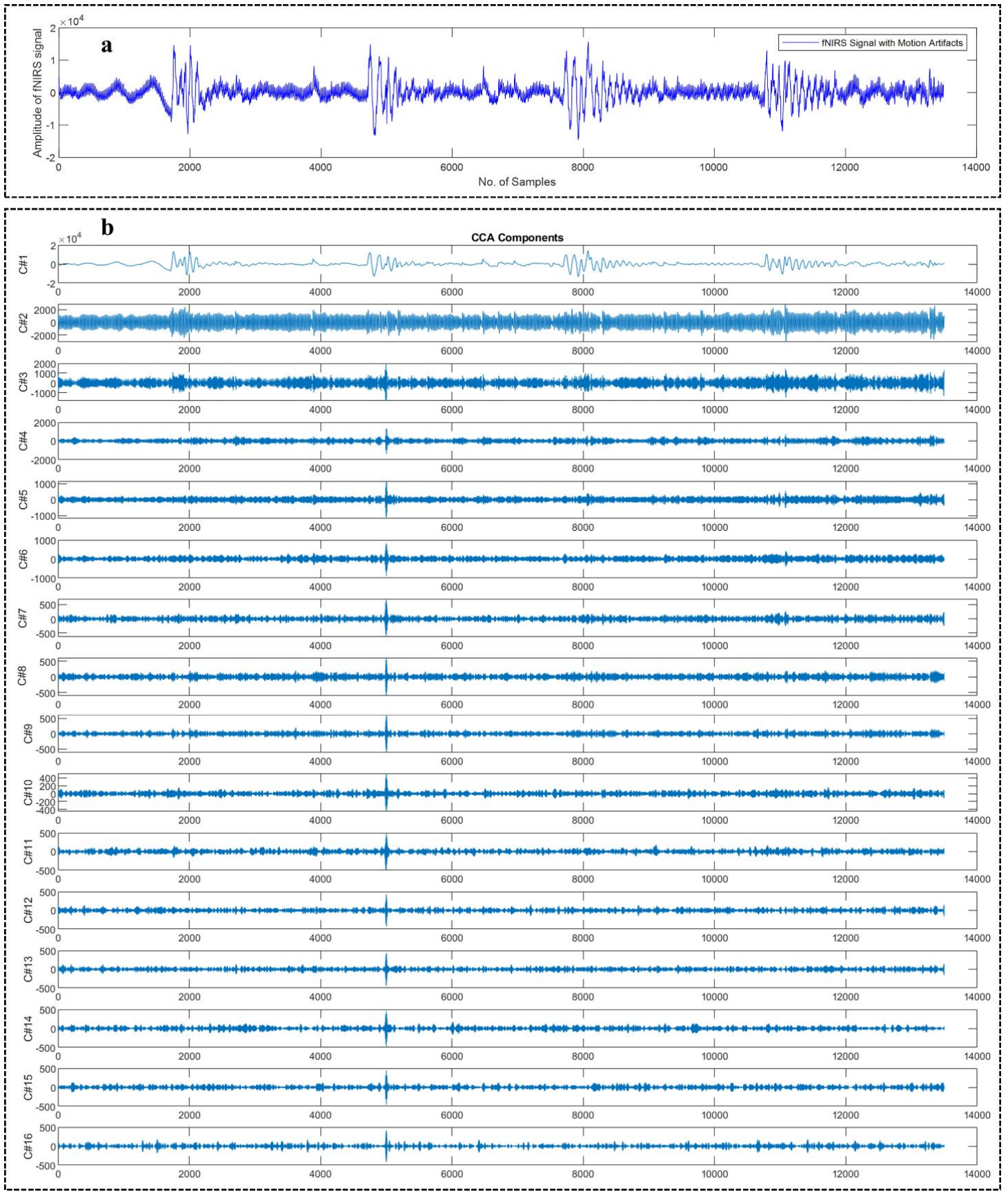

Figure 6: An example motion-corrupted single-channel EEG signal (a) and corresponding 16 CCA components generated from CCA algorithm (b)

## 3.5 Performance Metrics:

The efficacy and performance of each proposed artifact removal approach can be computed using the provided reference "ground truth" signal for each modality, as detailed before. Since the objective of each proposed technique is to reduce artifacts from the motion-artifact contaminated signal, calculating ΔSNR and percentage reduction in motion artifacts can assess the efficacy of that corresponding technique's capacity to remove artifacts. Hence, the difference in SNR before and after artifact removal (ΔSNR), and the improvement in correlation between motion corrupted and reference "ground truth" signals, expressed by the percentage reduction in motion artifact $\eta$ [33], are utilized as performance metrics.

For the calculation of ΔSNR, the following formula is used which was given in [33]:

$$\Delta SNR = 10 \ log_{10}\left(\frac{\sigma_x^2}{\sigma_{e_{after}}^2}\right) - 10 \ log_{10}\left(\frac{\sigma_x^2}{\sigma_{e_{before}}^2}\right) \tag{14}$$

where $\sigma_x^2$, $\sigma_{e_{before}}^2$, and $\sigma_{e_{after}}^2$ represent the variance of the reference "ground truth", motion corrupted signal, and cleaned signal, respectively.

To calculate the percentage reduction in motion artifact $\eta$, the following formula is used [33]:

$$\eta = 100 \ (1 - \frac{\rho_{clean} - \rho_{after}}{\rho_{clean} - \rho_{before}}) \tag{15}$$

Where $\rho_{before}$ is the correlation coefficient between the reference "ground truth" and motion-corrupted signals. The correlation coefficient between the reference "ground truth" and the cleaned signals is denoted by $\rho_{after}$ whereas $\rho_{clean}$ is the correlation between reference "ground truth" and motion corrupted signals over the epochs where motion artifact is absent.

In this study, we considered $\rho_{clean} = 1$ as in an ideal situation, the "reference ground truth" and the motion corrupted signal over the artifacts-free epochs would always be completely correlated. Hence, the following equation was used to estimate $\eta$:

$$\eta = 100 \ (1 - \frac{1 - \rho_{after}}{1 - \rho_{before}}) \tag{16}$$

## 4. Results and Discussion

The results obtained in this work, using the various novel artifact removal techniques are mentioned below where the performance metrics were calculated using Equations 14 and 16.

### 4.1 Motion Artifact Correction from EEG data:

All the algorithms (18 in total) were applied on all the 23 recordings of EEG. Figure 7a, Figure 7b, Figure 7c, and Figure 7d depicts 4 different examples of EEG recordings after the correction of the motion artifact using WPD(db2), WPD(db3), WPD(fk6), and WPD(fk8) methods, respectively whereas Figure 8a, and Figure 8b illustrate example EEG signals after the motion artifact correction using WPD(db1)-CCA, and WPD(fk4)-CCA techniques, respectively.

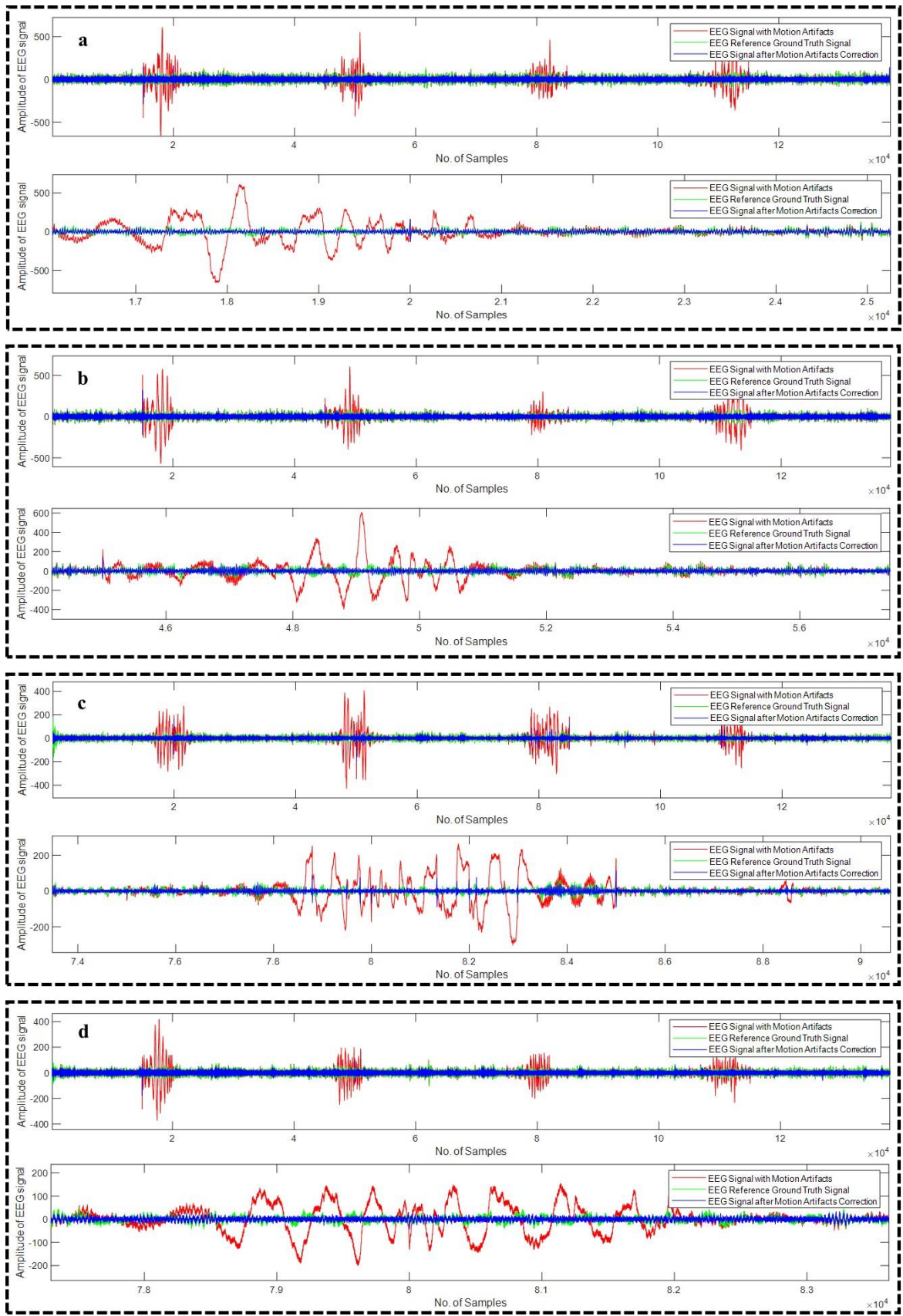

Figure 7: Motion artifact correction from different example EEG signals using WPD$_{(db2)}$(a), WPD$_{(db3)}$ (b), WPD$_{(fk6)}$ (c), and WPD$_{(fk8)}$ (d) techniques.

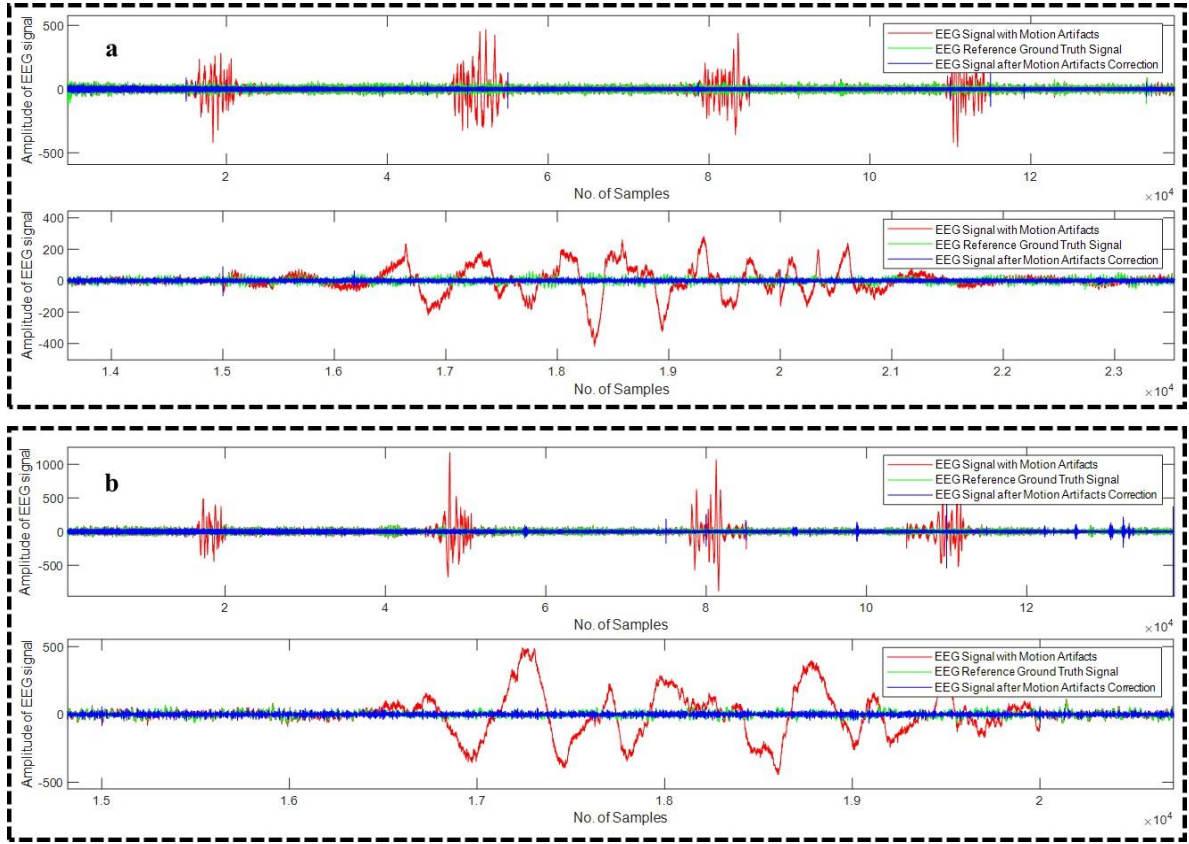

Figure 8: Motion artifact from example EEG signals using WPD$_{(db1)}$-CCA (a), and WPD$_{(fk4)}$-CCA (b) techniques

WPD: Among all the 12 different approaches (WPD$_{(db1)}$, WPD$_{(db2)}$, WPD$_{(db3)}$, WPD$_{(sym4)}$, WPD$_{(sym5)}$, WPD$_{(sym6)}$, WPD$_{(coif1)}$, WPD$_{(coif2)}$, WPD$_{(coif3)}$, WPD$_{(fk6)}$, WPD$_{(fk6)}$, and WPD$_{(fk8)}$), the highest average ΔSNR of 29.44 dB with a standard deviation of 9.93 was found when WPD$_{(db2)}$ algorithm was employed over all (23) EEG recordings. The best average percentage reduction in artifact was provided by WPD$_{(db1)}$ algorithm (53.48 %) among these 12 single-channel motion artifact correction techniques.

WPD-CCA: Six different approaches namely WPD$_{(db1)}$-CCA, WPD$_{(db2)}$-CCA, WPD$_{(db3)}$-CCA, WPD$_{(fk4)}$-CCA, WPD$_{(fk6)}$-CCA, and WPD$_{(fk8)}$-CCA were investigated all of which are two-stage motion artifacts correction techniques. The best average ΔSNR was found to be 30.76 dB when WPD$_{(db1)}$-CCA technique was applied over all the EEG records. The highest average percentage reduction in artifact was also provided by the same algorithm which is 59.51% among these 6 single-channel motion artifact correction techniques for EEG modality.

**4.2 Motion Artifact Correction from fNIRS data:**

All the algorithms (18 in total) were applied on all the 16 recordings of the fNIRS modality. Figure 9a, Figure 9b, Figure 9c, and Figure 9d depicts 4 different example fNIRS signals after the correction of the motion artifact using WPD$_{(sym5)}$, WPD$_{(sym6)}$, WPD$_{(coif2)}$, and WPD$_{(coif1)}$ techniques, respectively whereas Figure 10a, and Figure 10b illustrate example fNIRS signals after the motion artifact correction using WPD$_{(db1)}$-CCA, and WPD$_{(fk4)}$-CCA techniques, respectively.

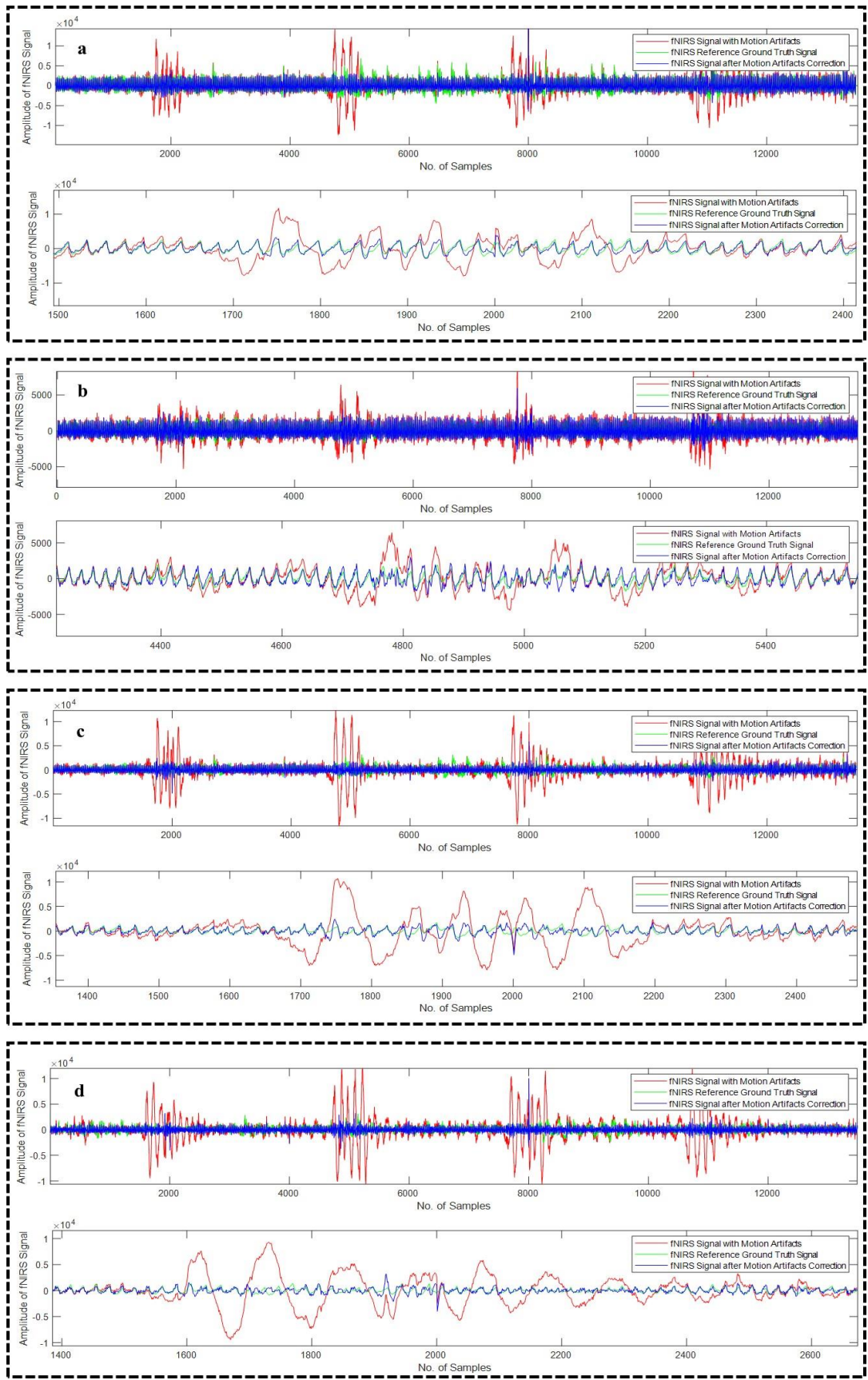

Figure 9: Motion artifact correction from example fNIRS signals using WPD$_{(sym5)}$ (a), WPD$_{(sym6)}$ (b), WPD$_{(coif2)}$ (c), and WPD$_{(coif1)}$ (d) techniques

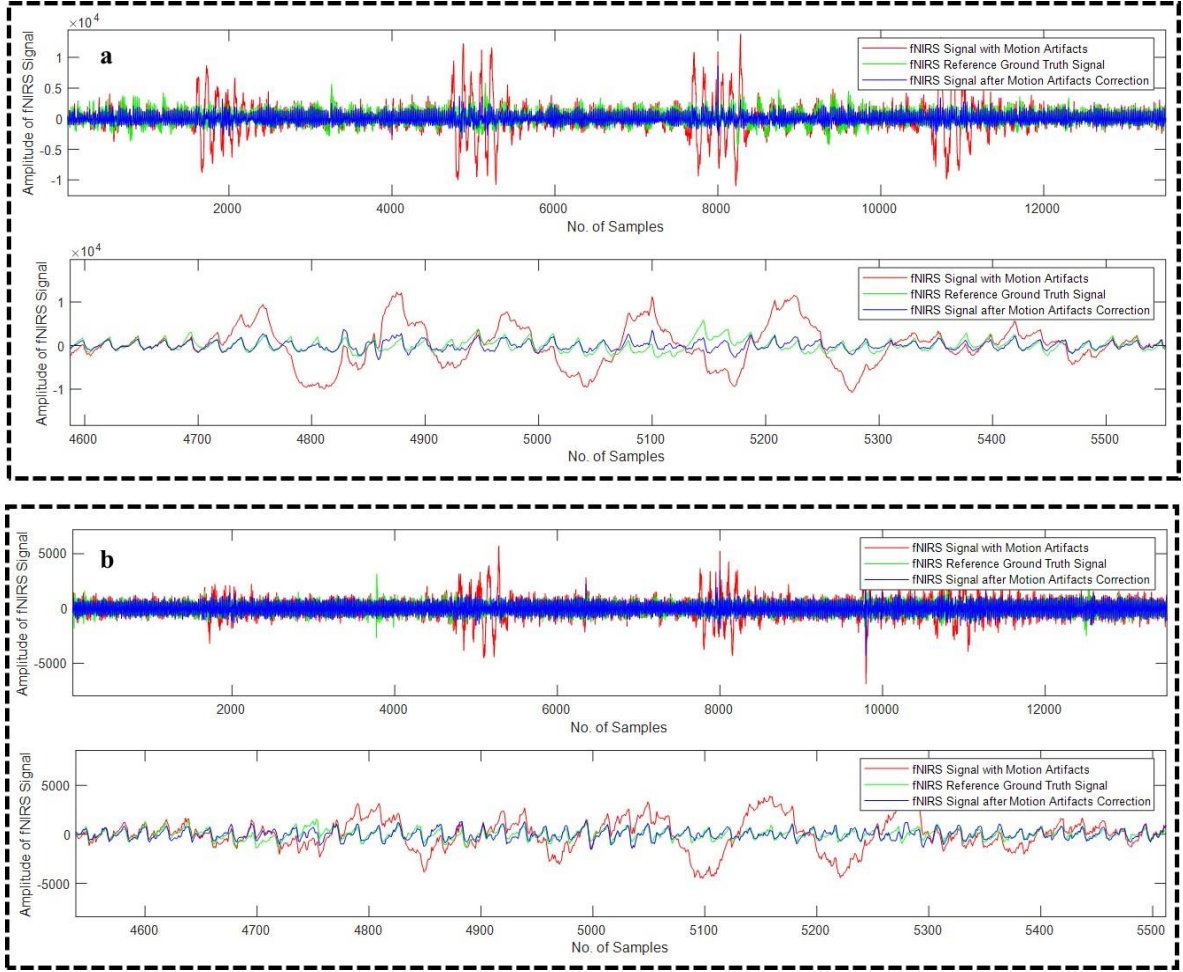

Figure 10: Motion artifact correction from example fNIRS signals using WPD$_{(db1)}$-CCA (a), and WPD$_{(fk4)}$-CCA (b) techniques

<u>WPD:</u> Among all the 12 different approaches (WPD$_{(db1)}$, WPD$_{(db2)}$, WPD$_{(db3)}$, WPD$_{(sym4)}$, WPD$_{(sym5)}$, WPD$_{(sym6)}$, WPD$_{(coif1)}$, WPD$_{(coif2)}$, WPD$_{(coif3)}$, WPD$_{(fk6)}$, WPD$_{(fk6)}$, and WPD$_{(fk8)}$), the highest average ΔSNR of 16.03 dB with a standard deviation of 4.31 was found when WPD$_{(db1)}$ algorithm was employed over all (16) fNIRS recordings. The best average percentage reduction in artifact was provided by WPD$_{(fk4)}$ algorithm among these 12 single-channel motion artifact correction techniques.

<u>WPD-CCA:</u> Finally, the six different approaches namely WPD$_{(db1)}$-CCA, WPD$_{(db2)}$-CCA, WPD$_{(db3)}$-CCA, WPD$_{(fk4)}$-CCA, WPD$_{(fk6)}$-CCA, and WPD$_{(fk8)}$-CCA, all of which are two-stage motion artifacts correction techniques, were investigated for fNIRS modality. The best average ΔSNR was found to be 16.55 dB when WPD$_{(db1)}$-CCA technique was applied over all the 16 fNIRS records. The highest average percentage reduction in artifact (41.40%) was provided by WPD$_{(fk8)}$-CCA technique among these 6 single-channel motion artifact correction techniques for fNIRS modality.

Table *1* summarizes the results obtained (average ΔSNR and average percentage reduction in motion artifacts $\eta$) using the artifact removal techniques proposed in this paper i.e. WPD$_{(db1)}$, WPD$_{(db2)}$, WPD$_{(db3)}$, WPD$_{(sym4)}$, WPD$_{(sym5)}$, WPD$_{(sym6)}$, WPD$_{(coif1)}$, WPD$_{(coif2)}$, WPD$_{(coif3)}$, WPD$_{(fk6)}$, WPD$_{(fk6)}$, WPD$_{(fk8)}$, WPD$_{(db1)}$-CCA, WPD$_{(db2)}$-CCA, WPD$_{(db3)}$-CCA, WPD$_{(fk4)}$-CCA, WPD$_{(fk6)}$-CCA, and WPD$_{(fk8)}$-CCA for all the EEG (23) and fNIRS (16) recordings. The values inside first brackets in

Table *1* denote the corresponding standard deviations.

Table 1: Average ΔSNR and average percentage reduction in artifacts (η) for all the EEG and fNIRS recordings. Corresponding standard deviations are shown inside the bracket. (*) represents the best-performing metrics.

| Type | Technique | EEG (23 records) | | fNIRS (16 records) | |
|---|---|---|---|---|---|
| | | Average ΔSNR (in dB) | Average η (in %) | Average ΔSNR (in dB) | Average η (in %) |
| Single-stage motion artifact correction techniques | WPD$_{(db1)}$ | 29.26 (10.29) | 53.48 (33.35)* | 16.03 (4.31) | 26.21 (26.38) |
| | WPD$_{(db2)}$ | 29.44 (9.93)* | 51.40 (33.59) | 15.99 (4.49) | 25.92 (28.86) |
| | WPD$_{(db3)}$ | 29.37 (10.01) | 50.74 (33.55) | 15.71 (4.52) | 26.05 (29.11) |
| | WPD$_{(sym4)}$ | 29.27 (10.05) | 50.40 (33.50) | 15.54 (4.55) | 26.14 (29.18) |
| | WPD$_{(sym5)}$ | 29.19 (10.09) | 50.20 (33.47) | 15.43 (4.57) | 26.17 (29.22) |
| | WPD$_{(sym6)}$ | 29.11 (10.12) | 50.05 (33.43) | 15.35 (4.59) | 26.16 (29.24) |
| | WPD$_{(coif1)}$ | 29.43 (9.94) | 51.34 (33.59) | 15.97 (4.49) | 25.94 (28.88) |
| | WPD$_{(coif2)}$ | 29.25 (10.06) | 50.35 (33.49) | 15.51 (4.56) | 26.15 (29.19) |
| | WPD$_{(coif3)}$ | 29.08 (10.13) | 50.00 (33.42) | 15.33 (4.60) | 26.15 (29.25) |
| | WPD$_{(fk4)}$ | 29.21 (9.87) | 52.58 (33.48) | 16.11 (4.42)* | 26.40 (27.53)* |
| | WPD$_{(fk6)}$ | 29.32 (10.03) | 50.55 (33.51) | 15.59 (4.54) | 26.20 (29.08) |
| | WPD$_{(fk8)}$ | 29.15 (10.10) | 50.15 (33.45) | 15.38 (4.58) | 26.25 (29.18) |
| Two-stage motion artifact correction techniques | WPD$_{(db1)}$-CCA | 30.76 (12.29)* | 59.51(25.99)* | 16.55 (6.29)* | 36.58 (11.22) |
| | WPD$_{(db2)}$-CCA | 30.35 (12.50) | 57.57 (25.89) | 14.50 (5.85) | 39.62 (10.59) |
| | WPD$_{(db3)}$-CCA | 29.42 (12.57) | 56.52 (25.71) | 13.72 (5.82) | 40.39 (10.60) |
| | WPD$_{(fk4)}$-CCA | 30.36 (12.65) | 58.83 (25.93) | 14.97 (6.25) | 38.32 (10.90) |
| | WPD$_{(fk6)}$-CCA | 29.12 (13.00) | 56.81 (25.16) | 13.81 (5.70) | 40.48 (10.43) |
| | WPD$_{(fk8)}$-CCA | 28.86 (12.77) | 55.88 (25.10) | 12.41 (5.51) | 41.40 (10.08)* |

It is evident from the results of

Table *1* that the cleaner EEG signals reconstructed using the WPD$_{(db1)}$ technique provided the highest average η value (53.48%, with corresponding ΔSNR value of 29.26 dB) compared to the other 11 types of single-stage motion artifact correction approaches whereas the greatest average ΔSNR value (29.44 dB) was provided by WPD$_{(db2)}$ with corresponding average η value of 51.40%. Among these 12 different single-stage artifact removal approaches, the lowest average $\eta$ (50.00%) and smallest ΔSNR (29.08 dB) was produced by WPD$_{(coif3)}$ method. When two-stage motion artifacts removal techniques were employed (WPD-CCA) using six different wavelet packets separately, the best average correlation improvement (59.51%) and best average ΔSNR value (30.76 dB) was produced by WPD$_{(db1)}$-CCA approach whereas the lowest performance was recorded utilizing WPD$_{(fk8)}$-CCA technique (average ΔSNR and $\eta$ values of 28.86 dB and 55.88%, respectively). Overall, An increase of 11.28% in the average percentage reduction in motion artifacts was found while the best-performing two-stage WPD$_{(db1)}$-CCA was incorporated compared to the best-performing single-stage motion artifact correction technique namely WPD$_{(db1)}$. Also, the average ΔSNR value improved by 4.48% (from 29.44 dB to 30.76 dB) while the best performing two-stage WPD$_{(db1)}$-CCA technique was utilized instead of the best-performing single-stage WPD$_{(db2)}$ method for the correction of motion artifacts from single-channel EEG recordings.

From

Table 1, the cleaner fNIRS signals reconstructed using WPD$_{(fk4)}$ technique provided the highest average $\eta$ value (26.40 %) compared to the other 11 types of single-stage motion artifact correction approaches. The greatest average ΔSNR value (16.11 dB) was also provided by the same approach. Among these 12 different single-stage artifact removal approaches, the lowest average $\eta$ (25.92%) was produced by WPD$_{(db2)}$ whereas the smallest ΔSNR value (15.33 dB) was produced by WPD$_{(coif3)}$. When two-stage motion artifacts removal techniques were employed (WPD-CCA) using six different wavelet packets for all the fNIRS signals, the best average correlation improvement (41.40%) was produced by WPD$_{(fk8)}$-CCA technique and the lowest average percentage reduction in artifacts (36.58 %) was generated from WPD$_{(db1)}$-CCA. On the other hand, the best average ΔSNR value (16.55 dB) was obtained from WPD$_{(db1)}$-CCA technique, and the WPD$_{(fk8)}$-CCA produced the lowest ΔSNR value of 12.41 dB. Overall, an increase of 56.82% in percentage reduction in motion artifacts was found while the best performing two-stage motion artifacts technique i.e. WPD$_{(fk8)}$-CCA was incorporated compared to the best performing single-stage motion artifact correction technique namely WPD$_{(fk4)}$. Also, an increase of 2.73% in ΔSNR value was found when best performing two-stage WPD$_{(db1)}$-CCA was employed instead of the best-performing single-stage WPD$_{(fk4)}$ technique.

From Table 1, it is clear that two-stage artifacts correction techniques performed relatively better compared to the single-stage artifacts correction approaches for both EEG and fNIRS modalities.

Authors of [37] found that no brain activity was registered in trials 12 and 15. Moreover, they found a poor correlation coefficient over the clean epochs of the recordings of 12 and 15, and hence, they carried out their investigation on the remaining 21 recordings of EEG. We have also observed a similar situation in this work. Trials 12 and 15 consistently produced very bad performance metrics (ΔSNR and $\eta$ values) while both single-stage and two-stage artifact reduction techniques were applied proposed in this paper.

The authors of this work are presenting a 2nd table (**Error! Reference source not found.**) that illustrates the average ΔSNR and average percent reduction in motion artifacts using WPD$_{(db1)}$, WPD$_{(sym4)}$, WPD$_{(coif1)}$, and WPD$_{(fk4)}$. This time the faulty trials (trials 12 and 15) were excluded and the experiments were conducted on the remaining 21 sets of EEG recordings. The motion corrupted signal was decomposed into 16 sub-band components using WPD and then the cleaner signals were generated by simply discarding the lowest-frequency approximation sub-band component (S15, Figure 7) and adding the remaining 15 sub-band components directly. During this process, the reference ground truth signal was only used to compute the performance metrics.

Table 2: Average ΔSNR and average percentage reduction in artifacts ($\eta$) for 21 recordings of EEG modality. Corresponding standard deviations are shown inside the first bracket. (*) denotes the best-performing metrics.

| Type | Method | EEG (21 records) |
|---|---|---|

|  |  | **Average ΔSNR (in dB)** | **Average $\eta$ (in %)** |
|---|---|---|---|
| Single-stage motion artifact correction techniques | WPD(db1) | 26.20 (6.35) | 60.22 (21.79)* |
|  | WPD(sym4) | 26.46 (6.56) | 57.23 (22.11) |
|  | WPD(coif1) | 26.70 (6.54)* | 58.19 (22.04) |
|  | WPD(fk4) | 26.36 (6.36) | 59.37 (21.90) |

From table 2, it is clear that the cleaner EEG signals reconstructed using WPD(db1) technique provided the highest average $\eta$ value (60.22 %, corresponding ΔSNR value of 26.20 dB) compared to the other 3 types of single-stage motion artifact correction approaches whereas the greatest average ΔSNR value (26.70 dB) was produced by WPD(coif1) with an average $\eta$ value of 58.19 %. The values obtained following this process is a clear indication that without the availability of "reference ground truth signal", correction of motion artifacts from EEG signal is still possible. The similar approach can be used for motion artifacts correction from fNIRS signals also, but left as a future work.

## 5. Discussion

In this extensive work, we have proposed two novel methods (WPD and WPD-CCA) using four different wavelet packet families with three different vanishing moments, resulting in 18 different techniques (WPD(db1), WPD(db2), WPD(db3), WPD(sym4), WPD(sym5), WPD(sym6), WPD(coif1), WPD(coif2), WPD(coif3), WPD(fk6), WPD(fk6), WPD(fk8), WPD(db1)-CCA, WPD(db2)-CCA, WPD(db3)-CCA, WPD(fk4)-CCA, WPD(fk6)-CCA, and WPD(fk8)-CCA) for the correction of motion artifacts from single-channel EEG and fNIRS recordings. The performance metrics (ΔSNR and $\eta$) calculated and reported in the "Results" section utilizing these 18 approaches are a clear indication of the efficacy of our proposed techniques. Both the Daubechies and Fejer-Korovkin wavelet packet families relatively performed better compared to the Symlet and Coiflet wavelet packet families in removing motion artifacts from EEG and fNIRS recordings. For this reason, while implementing the two-stage artifacts correction technique, we have used only the Daubechies and Fejer-Korovkin wavelet packet families.

As previously stated, DWT, EMD, EEMD, VMD, EMD-ICA, EMD-CCA, EEMD-ICA, EEMD-CCA, VMD-CCA, SSA, DWT along with approximation sub-band filtering, adaptive filtering (ARX model with exogenous input), etc. were commonly employed for the correction of movement artifacts from motion corrupted EEG and fNIRS signals. Each of these methods suffers from some limitations.

Using DWT-based approaches, to improve signal quality from motion-corrupted physiological data, selecting the suitable wavelet is critical and rather complex. To date, there is no hard and fast rule for selecting the appropriate wavelet for the specific physiological signal of interest; instead, wavelets are often selected depending on the morphology of the signal. As a result, improper wavelet selection would result in inefficient denoising.

The EMD-based motion artifact reduction approach suffers heavily from the "mode mixing" issue [33], which may result in an incorrect outcome. To fix this problem, the EEMD approaches are employed [33, 36]. Although EEMD is not affected by the mode mixing problem, it still requires a prior declaration of the number of ensembles to be employed which is determined through trial and error basis [33].

To make use of the SSA algorithm, for the correction of movement artifacts from physiological signals, a prior declaration of the window length and the required number of reconstruction components is necessary, which makes SSA inefficient as well [37].

The authors of [40] employed DWT along with approximation sub-band filtering using total variation (TV) and weighted TV. While reconstructing the cleaner signal, the first three high-frequency detailed sub-band signals were rejected since they included no important information from the EEG signal. But, detecting non-useful sub-band signals when utilizing DWT-based algorithms is very challenging for removing motion artifacts from EEG and fNIRS signals. Furthermore, the value of the regularization factor used to address the optimization problem of TV and MTV approaches was picked without explanation.

In [52], they studied the autoregressive exogenous input model (adaptive technique) to model motion corrupted segments as output and IMU data as exogenous input. Only four test participants' fNIRS data were used by the authors to demonstrate the efficacy of their prescribed approach. One of the most important aspects of adopting this technique is the precise synchronization of fNIRS and IMU data. Furthermore, if the epoch duration of the motion artifacts is sufficiently long (specifically, the sample size), modeling the artifacts mathematically using the least square method would necessitate higher-order models which would eventually cause instability. Hence, incorporating this method to remove motion artifacts would be extremely difficult in a real-world scenario.

ICA and CCA algorithms are multi-channel signal processing algorithms meaning there must be two (or more) channel data as input. Therefore, ICA and CCA algorithms can not be incorporated independently for the processing of single-channel data. Also, since ICA uses higher-order statistics (HOS) and CCA uses second-order statistics (SOS) {Sweeney, 2012 #66}, the CCA algorithm is computationally efficient in comparison with ICA. That is why previous studies as well as this study used the CCA algorithm as a 2$^{nd}$ stage signal processing method.

WPD is the more generalized version of DWT but the former provides better signal decomposition which enhances the signal quality for further processing. Also, WPD is better in denoising in the sense that there is no necessity of identifying and discarding any sub-band signals other than the motion corrupted sub-band component. Also, the results obtained in this work utilizing the WPD method for 12 different wavelet packets, show a little variation while computing $\Delta SNR$ and $\eta$. This is a clear indication that, applying WPD compared to the DWT is much more robust and efficient in terms of performance metrics improvement.

Although the two-stage motion artifacts removal approaches (WPD-CCA) proposed in this paper performed better compared to the single-stage artifacts correction techniques using WPD, the WPD-CCA technique will not be able to identify the motion corrupted CCA components In the absence of a ground truth signal, which is a limitation of two-stage artifacts removal technique. Hassan *et al.* provided an alternate technique in [65], in which the authors employed the autocorrelation function to detect the motion corrupted components. The automated artifact component selection approach introduced in [65] employing the autocorrelation function has not been experimented in this study and left as a future study.

However, even in the absence of the "reference ground truth" signal, our proposed single-stage motion artifact reduction approach (WPD) would produce optimal results. While decomposing the signal of interest (EEG/fNIRS) using WPD, it was visually seen that the approximation sub-band component (having the lowest frequency band compared to the rest of the sub-band components) included the highest percentage of motion artifacts. Hence, discarding this noisy sub-band component and reconstruction of the signal using the remaining sub-band signals would reduce the motion artifacts to a great extent. The validation of this statement is supported by Table 2 where the performance metrics ($\Delta SNR$ and $\eta$) were reported and produced acceptable results.

Throughout this work, while estimating the percentage reduction in motion artifacts $\eta$, we have considered Equation 16, instead of 15. Where we have assumed that $\rho_{clean} = 1$ as in an ideal situation, the "reference ground truth" and the motion corrupted signal over the artifacts-free epochs would always be completely correlated. But in practice, the value of $\rho_{clean}$ would always be less than 1. Because it is impossible to extract a "reference ground truth" signal which would completely be similar while compared with a motion-corrupted signal during the artifacts-free epochs. It is counter-intuitive that lower value of $\rho_{clean}$ would produce a lesser value of $\eta$. Rather it is just the opposite. For example, let $\rho_{before} = 0.6$; $\rho_{after} = 0.8$; $\rho_{clean} = 0.95$, from Equation 15, we would get $\eta$ equals 57.14% and Equation 16 would give 50%. That is why choosing $\rho_{clean} = 1$ would give a worst-case scenario result. Also, this same formula is used in [40-42] assuming the ideal "reference ground truth signal".

## 6. Conclusions

In this extensive study, two novel motion artifact removal techniques have been proposed, namely wavelet packet decomposition (WPD), and WPD in combination with canonical correlation analysis (WPD-CCA) for EEG and fNIRS modalities. Further, the proposed algorithms were investigated by 18 different approaches where four different wavelet packet families namely Daubechies, Symlet, Coiflet, and Fejer-Korovkin wavelet packet families were utilized. WPD-CCA techniques can be used on single-channel recordings as the WPD algorithm can decompose a single-channel signal into a predefined number of sub-band components which can be fed as the input channels for the CCA algorithm. The performance parameters obtained from all these approaches are a clear indication of the efficacy of these algorithms. The novel WPD$_{(db1)}$-CCA and WPD$_{(fk8)}$-CCA technique provided the best performance in terms of the percentage reduction in motion artifacts (59.51% and 41.40%) when analyzing the EEG and fNIRS data, respectively. On the other hand, the WPD$_{(db1)}$-CCA technique generated the highest average ΔSNR (30.76 dB and 16.55 dB) for both EEG and fNIRS signals. An alternative approach for removing motion artifacts from EEG signals using the WPD method has also been proposed where the lowest-frequency approximation sub-band component was discarded and clean EEG signal was reconstructed by adding up the remaining sub-band components. By computing the performance metrics, it has been shown that this single-stage motion artifacts correction technique is also capable of removing motion artifacts to a great extent. In the future, deep learning-based models will be investigated for the automated detection and removal of artifacts in physiological signals (EEG, ECG, EMG, PPG, fNIRS, etc.). New methods based on the use of different multivariate signal processing approaches will be developed for the elimination of other artifacts from the EEG and fNIRS signals that are recorded using multiple electrodes.

**Fundings:**
This work was made possible by Qatar National Research Fund (QNRF) NPRP12S-0227-190164 and International Research Collaboration Co-Fund (IRCC) grant: IRCC-2021-001 and Universiti Kebangsaan Malaysia (UKM) under Grant GUP-2021-019, Grant TAP-K017701, and DPK-2021-001. The statements made herein are solely the responsibility of the authors.

**Author Contributions**
Conceptualization, Md Shafayet Hossain, Muhammad E. H. Chowdhury, Mamun Bin Ibne Reaz, Sawal Ali, Ahmad Bakar, Serkan Kiranyaz, Amith Khandakar, Mohammed Alhatou, Rumana Habib and Muhammad Maqsud Hossain; Data curation, Md Shafayet Hossain and Mamun Bin Ibne Reaz; Formal analysis, Md Shafayet Hossain; Funding acquisition, Muhammad E. H. Chowdhury and Mamun Bin Ibne Reaz; Methodology, Md Shafayet Hossain, Muhammad E. H. Chowdhury, Mamun Bin Ibne Reaz, Sawal Ali, Amith Khandakar and Rumana Habib; Project administration, Muhammad E. H. Chowdhury and Mamun Bin Ibne Reaz; Resources, Muhammad E. H. Chowdhury; Software, Muhammad E. H. Chowdhury; Supervision, Muhammad E. H. Chowdhury and Mamun Bin Ibne Reaz; Validation, Md Shafayet Hossain; Visualization, Md Shafayet Hossain; Writing – original draft, Md Shafayet Hossain, Muhammad E. H. Chowdhury, Mamun Bin Ibne Reaz, Sawal Ali, Ahmad Bakar, Serkan Kiranyaz, Amith Khandakar, Mohammed Alhatou, Rumana Habib and Muhammad Maqsud Hossain; Writing – review & editing, Md Shafayet Hossain, Muhammad E. H. Chowdhury, Mamun Bin Ibne Reaz and Muhammad Maqsud Hossain.

**Acknowledgments**
The dataset used in this experiment is kindly shared in the PhysioNet database by Sweeney *et al.* [32, 33, 63].

**Ethical Statement**

No ethical statement is to be declared. The dataset used in this study is publicly available in the PhysioNet database and the authors of this study did not collect the dataset. Sweeney *et al.* [32, 33, 63] collected this dataset with ethical approval.

**Conflicts of Interest**

The authors declare no conflict of interest.

**References**

[1] J. C. Henry, "Electroencephalography: basic principles, clinical applications, and related fields," *Neurology,* vol. 67, no. 11, pp. 2092-2092-a, 2006.

[2] M. Nuwer, "Assessment of digital EEG, quantitative EEG, and EEG brain mapping: report of the American Academy of Neurology and the American Clinical Neurophysiology Society," *Neurology,* vol. 49, no. 1, pp. 277-292, 1997.

[3] A. Shoeb, J. Guttag, S. Schachter, D. Schomer, B. Bourgeois, and S. T. Treves, "Detecting seizure onset in the ambulatory setting: Demonstrating feasibility," in *2005 IEEE Engineering in Medicine and Biology 27th Annual Conference*, 2006, pp. 3546-3550: IEEE.

[4] R. R. Sharma, P. Varshney, R. B. Pachori, and S. K. Vishvakarma, "Automated system for epileptic EEG detection using iterative filtering," *IEEE Sensors Letters,* vol. 2, no. 4, pp. 1-4, 2018.

[5] C. Berka *et al.*, "EEG quantification of alertness: methods for early identification of individuals most susceptible to sleep deprivation," in *Biomonitoring for Physiological and Cognitive Performance during Military Operations*, 2005, vol. 5797, pp. 78-89: International Society for Optics and Photonics.

[6] C. Berka *et al.*, "Real-time analysis of EEG indexes of alertness, cognition, and memory acquired with a wireless EEG headset," *International Journal of Human-Computer Interaction,* vol. 17, no. 2, pp. 151-170, 2004.

[7] C. Papadelis *et al.*, "Indicators of sleepiness in an ambulatory EEG study of night driving," in *2006 International Conference of the IEEE Engineering in Medicine and Biology Society*, 2006, pp. 6201-6204: IEEE.

[8] R. Tripathy and U. R. Acharya, "Use of features from RR-time series and EEG signals for automated classification of sleep stages in deep neural network framework," *Biocybernetics and Biomedical Engineering,* vol. 38, no. 4, pp. 890-902, 2018.

[9] V. Gupta, M. D. Chopda, and R. B. Pachori, "Cross-subject emotion recognition using flexible analytic wavelet transform from EEG signals," *IEEE Sensors Journal,* vol. 19, no. 6, pp. 2266-2274, 2018.

[10] R. Stevens, T. Galloway, C. Berka, and Future, "Integrating EEG models of cognitive load with machine learning models of scientific problem solving," *Augmented Cognition: Past, Present and Future,* vol. 2, pp. 55-65, 2006.

[11] C. J. Bell, P. Shenoy, R. Chalodhorn, and R. P. Rao, "Control of a humanoid robot by a noninvasive brain–computer interface in humans," *Journal of neural engineering,* vol. 5, no. 2, p. 214, 2008.

[12] J. C. Lee and D. S. Tan, "Using a low-cost electroencephalograph for task classification in HCI research," in *Proceedings of the 19th annual ACM symposium on User interface software and technology*, 2006, pp. 81-90.

[13] T. J. Sullivan, S. R. Deiss, T.-P. Jung, and G. Cauwenberghs, "A brain-machine interface using dry-contact, low-noise EEG sensors," in *2008 IEEE International Symposium on Circuits and Systems*, 2008, pp. 1986-1989: IEEE.

[14] J. R. Wolpaw and D. J. McFarland, "Control of a two-dimensional movement signal by a noninvasive brain-computer interface in humans," *Proceedings of the national academy of sciences,* vol. 101, no. 51, pp. 17849-17854, 2004.

[15] P. Gaur, R. B. Pachori, H. Wang, and G. Prasad, "A multi-class EEG-based BCI classification using multivariate empirical mode decomposition based filtering and Riemannian geometry," *Expert Systems with Applications,* vol. 95, pp. 201-211, 2018.

[16] P. Gaur, R. B. Pachori, H. Wang, and G. Prasad, "An automatic subject specific intrinsic mode function selection for enhancing two-class EEG-based motor imagery-brain computer interface," *IEEE Sensors Journal,* vol. 19, no. 16, pp. 6938-6947, 2019.

[17] A. Rahman *et al.*, "Multimodal EEG and Keystroke Dynamics Based Biometric System Using Machine Learning Algorithms," *IEEE Access,* vol. 9, pp. 94625-94643, 2021.

[18] S. Sangani, A. Lamontagne, and J. Fung, "Cortical mechanisms underlying sensorimotor enhancement promoted by walking with haptic inputs in a virtual environment," *Progress in brain research,* vol. 218, pp. 313-330, 2015.

[19] S. C. Bunce, M. Izzetoglu, K. Izzetoglu, B. Onaral, and K. Pourrezaei, "Functional near-infrared spectroscopy," *IEEE engineering in medicine biology magazine,* vol. 25, no. 4, pp. 54-62, 2006.

[20] T. J. Huppert, S. G. Diamond, M. A. Franceschini, and D. A. Boas, "HomER: a review of time-series analysis methods for near-infrared spectroscopy of the brain," *Applied optics,* vol. 48, no. 10, pp. D280-D298, 2009.

[21] L. Holper, T. Muehlemann, F. Scholkmann, K. Eng, D. Kiper, and M. Wolf, "Testing the potential of a virtual reality neurorehabilitation system during performance of observation, imagery and imitation of motor actions recorded by wireless functional near-infrared spectroscopy (fNIRS)," *Journal of neuroengineering rehabilitation,* vol. 7, no. 1, pp. 1-13, 2010.

[22] K. Izzetoglu, S. Bunce, B. Onaral, K. Pourrezaei, and B. Chance, "Functional optical brain imaging using near-infrared during cognitive tasks," *International Journal of human-computer interaction,* vol. 17, no. 2, pp. 211-227, 2004.

[23] X. Cui, S. Bray, D. M. Bryant, G. H. Glover, and A. L. Reiss, "A quantitative comparison of NIRS and fMRI across multiple cognitive tasks," *Neuroimage,* vol. 54, no. 4, pp. 2808-2821, 2011.

[24] F. Matthews, B. A. Pearlmutter, T. E. Wards, C. Soraghan, and C. Markham, "Hemodynamics for brain-computer interfaces," *IEEE Signal Processing Magazine,* vol. 25, no. 1, pp. 87-94, 2007.

[25] M. J. Khan and K.-S. Hong, "Passive BCI based on drowsiness detection: an fNIRS study," *Biomedical optics express,* vol. 6, no. 10, pp. 4063-4078, 2015.

[26] K.-S. Hong, N. Naseer, and Y.-H. Kim, "Classification of prefrontal and motor cortex signals for three-class fNIRS–BCI," *Neuroscience letters,* vol. 587, pp. 87-92, 2015.

[27] M. E. Chowdhury, A. Khandakar, B. Hossain, and K. Alzoubi, "Effects of the phantom shape on the gradient artefact of electroencephalography (EEG) data in simultaneous EEG–fMRI," *Applied Sciences,* vol. 8, no. 10, p. 1969, 2018.

[28] M. E. Chowdhury *et al.*, "Reference layer artefact subtraction (RLAS): electromagnetic simulations," *IEEE Access,* vol. 7, pp. 17882-17895, 2019.

[29] M. E. Chowdhury, A. Khandakar, K. J. Mullinger, N. Al-Emadi, and R. Bowtell, "Simultaneous EEG-fMRI: evaluating the effect of the EEG cap-cabling configuration on the gradient artifact," *Frontiers in neuroscience,* vol. 13, p. 690, 2019.

[30] H.-D. Nguyen, S.-H. Yoo, M. R. Bhutta, and K.-S. Hong, "Adaptive filtering of physiological noises in fNIRS data," *Biomedical engineering online,* vol. 17, no. 1, pp. 1-23, 2018.

[31] M. K. Islam, A. Rastegarnia, and Z. Yang, "Methods for artifact detection and removal from scalp EEG: A review," *Neurophysiologie Clinique/Clinical Neurophysiology,* vol. 46, no. 4-5, pp. 287-305, 2016.

[32] K. T. Sweeney, T. E. Ward, and S. F. McLoone, "Artifact removal in physiological signals—Practices and possibilities," *IEEE transactions on information technology in biomedicine,* vol. 16, no. 3, pp. 488-500, 2012.

[33] K. T. Sweeney, S. F. McLoone, and T. E. Ward, "The use of ensemble empirical mode decomposition with canonical correlation analysis as a novel artifact removal technique," *IEEE transactions on biomedical engineering,* vol. 60, no. 1, pp. 97-105, 2012.

[34] A. N. Akansu, R. A. Haddad, P. A. Haddad, and P. R. Haddad, *Multiresolution signal decomposition: transforms, subbands, and wavelets*. Academic press, 2001.

[35] N. E. Huang *et al.*, "The empirical mode decomposition and the Hilbert spectrum for nonlinear and non-stationary time series analysis," *Proceedings of the Royal Society of London. Series A: mathematical, physical engineering sciences,* vol. 454, no. 1971, pp. 903-995, 1998.

[36] Z. Wu and N. E. Huang, "Ensemble empirical mode decomposition: a noise-assisted data analysis method," *Advances in adaptive data analysis,* vol. 1, no. 01, pp. 1-41, 2009.

[37] A. K. Maddirala and R. A. Shaik, "Motion artifact removal from single channel electroencephalogram signals using singular spectrum analysis," *Biomedical Signal Processing Control,* vol. 30, pp. 79-85, 2016.

[38] R. Vautard, P. Yiou, and M. Ghil, "Singular-spectrum analysis: A toolkit for short, noisy chaotic signals," *Physica D: Nonlinear Phenomena,* vol. 58, no. 1-4, pp. 95-126, 1992.

[39] P. S. Kumar, R. Arumuganathan, K. Sivakumar, and C. Vimal, "Removal of ocular artifacts in the EEG through wavelet transform without using an EOG reference channel," *Int. J. Open Problems Compt. Math,* vol. 1, no. 3, pp. 188-200, 2008.

[40] P. Gajbhiye, R. K. Tripathy, A. Bhattacharyya, and R. B. Pachori, "Novel approaches for the removal of motion artifact from EEG recordings," *IEEE Sensors Journal,* vol. 19, no. 22, pp. 10600-10608, 2019.

[41] P. Gajbhiye, N. Mingchinda, W. Chen, S. C. Mukhopadhyay, T. Wilaiprasitporn, and R. K. Tripathy, "Wavelet Domain Optimized Savitzky–Golay Filter for the Removal of Motion Artifacts From EEG Recordings," *IEEE Transactions on Instrumentation and Measurement,* vol. 70, pp. 1-11, 2020.

[42] M. S. Hossain *et al.*, "Motion Artifacts Correction from EEG and fNIRS Signals using Novel Multiresolution Analysis," *IEEE Access*, 2022.

[43] K. Dragomiretskiy and D. Zosso, "Variational mode decomposition," *IEEE transactions on signal processing,* vol. 62, no. 3, pp. 531-544, 2013.

[44] F. C. Robertson, T. S. Douglas, and E. M. Meintjes, "Motion artifact removal for functional near infrared spectroscopy: a comparison of methods," *IEEE Transactions on Biomedical Engineering,* vol. 57, no. 6, pp. 1377-1387, 2010.

[45] R. Cooper *et al.*, "A systematic comparison of motion artifact correction techniques for functional near-infrared spectroscopy," *Frontiers in neuroscience,* vol. 6, p. 147, 2012.

[46] S. Brigadoi *et al.*, "Motion artifacts in functional near-infrared spectroscopy: a comparison of motion correction techniques applied to real cognitive data," *Neuroimage,* vol. 85, pp. 181-191, 2014.

[47] K. T. Sweeney, H. Ayaz, T. E. Ward, M. Izzetoglu, S. F. McLoone, and B. Onaral, "A methodology for validating artifact removal techniques for physiological signals," *IEEE transactions on information technology in biomedicine,* vol. 16, no. 5, pp. 918-926, 2012.

[48] F. Scholkmann, S. Spichtig, T. Muehlemann, and M. Wolf, "How to detect and reduce movement artifacts in near-infrared imaging using moving standard deviation and spline interpolation," *Physiological measurement,* vol. 31, no. 5, p. 649, 2010.

[49] B. Molavi and G. A. Dumont, "Wavelet-based motion artifact removal for functional near-infrared spectroscopy," *Physiological measurement,* vol. 33, no. 2, p. 259, 2012.

[50] J. W. Barker, A. Aarabi, and T. J. Huppert, "Autoregressive model based algorithm for correcting motion and serially correlated errors in fNIRS," *Biomedical optics express,* vol. 4, no. 8, pp. 1366-1379, 2013.

[51] A. M. Chiarelli, E. L. Maclin, M. Fabiani, and G. Gratton, "A kurtosis-based wavelet algorithm for motion artifact correction of fNIRS data," *NeuroImage,* vol. 112, pp. 128-137, 2015.

[52] M. R. Siddiquee, J. S. Marquez, R. Atri, R. Ramon, R. Perry Mayrand, and O. Bai, "Movement artefact removal from NIRS signal using multi-channel IMU data," *Biomedical engineering online,* vol. 17, no. 1, pp. 1-16, 2018.

[53] S. Jahani, S. K. Setarehdan, D. A. Boas, and M. A. Yücel, "Motion artifact detection and correction in functional near-infrared spectroscopy: a new hybrid method based on spline interpolation method and Savitzky–Golay filtering," *Neurophotonics,* vol. 5, no. 1, p. 015003, 2018.

[54] S. Mallat, "A wavelet tour of signal processing (S. Mallat, Ed.)," ed: Academic Press, 2009.

[55] S. Jaffard, Y. Meyer, and R. D. Ryan, *Wavelets: tools for science and technology*. SIAM, 2001.

[56] O. Farooq and S. Datta, "Mel filter-like admissible wavelet packet structure for speech recognition," *IEEE signal processing letters,* vol. 8, no. 7, pp. 196-198, 2001.

[57] S. Sanei and J. A. Chambers, *EEG signal processing*. John Wiley & Sons, 2013.

[58] J. T. Gwin, K. Gramann, S. Makeig, and D. P. Ferris, "Removal of movement artifact from high-density EEG recorded during walking and running," *Journal of neurophysiology,* vol. 103, no. 6, pp. 3526-3534, 2010.

[59] H. Hotelling, "Relations between two sets of variates," in *Breakthroughs in statistics*: Springer, 1992, pp. 162-190.

[60] C. J. James and C. W. Hesse, "Independent component analysis for biomedical signals," *Physiological measurement,* vol. 26, no. 1, p. R15, 2004.

[61] W. De Clercq, A. Vergult, B. Vanrumste, W. Van Paesschen, and S. Van Huffel, "Canonical correlation analysis applied to remove muscle artifacts from the electroencephalogram," *IEEE transactions on Biomedical Engineering,* vol. 53, no. 12, pp. 2583-2587, 2006.

[62] M. Borga and H. Knutsson, "A canonical correlation approach to blind source separation," *Report LiU-IMT-EX-Department of Biomedical Engineering, Linkping University,* 2001.

[63] A. L. Goldberger *et al.*, "PhysioBank, PhysioToolkit, and PhysioNet: components of a new research resource for complex physiologic signals," *circulation,* vol. 101, no. 23, pp. e215-e220, 2000.

[64] M. G. Tsipouras, "Spectral information of EEG signals with respect to epilepsy classification," *EURASIP Journal on Advances in Signal Processing,* vol. 2019, no. 1, pp. 1-17, 2019.

[65] M. Hassan, S. Boudaoud, J. Terrien, B. Karlsson, and C. Marque, "Combination of canonical correlation analysis and empirical mode decomposition applied to denoising the labor electrohysterogram," *IEEE Transactions on Biomedical Engineering,* vol. 58, no. 9, pp. 2441-2447, 2011.